# Harnessing LLMs for Reliable Academic Supervision: A Comparative Study

**Akash Raj**

*M.Tech, School of Artificial Intelligence and Data Science*

*Indian Institute of Technology Jodhpur*

m25ai1116@iitj.ac.in

ORCID: 0009-0005-5666-3967

### Abstract

Large language models routinely produce fluent answers to single-shot prompts, yet deploying them as reliable components of a domain decision system is substantially harder. Closing this gap is the work of harness engineering: the deliberate composition of deterministic scaffolding (symbolic filters, retrieval, schema-typed I/O, LLM-as-judge loops, HITL gates, persistent state, audit trails) around an LLM core. We present a case study in academic supervision, a domain combining high stakes recommendation, longitudinal accountability, and structured operational workflows.

We compare baseline Academic Supervision Assistant (ASA), a GPT-5 chatbot with no scaffolding, against a multi-module system, Academic Supervision System (ASuS) that wraps the much smaller GPT-4o-mini in a LangGraph harness with symbolic-semantic retrieval, schema validated outputs, LLM-as-judge with bounded retry, HITL gates, deterministic weighted risk scoring with LLM narration, and a per node SQLite audit trail. The evaluation rubric is retargeted at six harness mechanism dimensions (grounding, explainability, consistency, process integrity, cognitive load and constraint adherence). A blind ten rater hybrid evaluation, supplemented by a 2 × 2 model harness ablation, finds that ASuS, despite using a much smaller base model, outscores ASA on every dimension. Across ten raters the pooled mean for ASuS is 4.08 versus 1.23 for ASA, and 8 of 10 raters reject the null at $\alpha = 0.05$ on a paired Wilcoxon test; full numbers are in sections 6.4 and 6.7. The ablation confirms that the structural contributions of the harness are largely model invariant. We extract seven recurring harness engineering patterns and argue that where reliability, traceability, and institutional consistency matter more than open ended fluency, harness engineering challenges the prevailing 'bigger model is better' intuition.

## 1. Introduction

The dominant narrative around large language models (LLMs) has emphasized scale, bigger models, longer context windows, more pre-training and post-training data [Bommasani et al. 2021]. Yet practitioners building domain systems repeatedly encounter the same gap: a model that solves a task in a Jupyter notebook does not, on its own, become a reliable production component. That production reliability is not free; it comes from engineering the components around the LLM, retrieving data from private corpora, validating outputs against a schema or a human, and maintaining audit trails so decisions can be defended. This scaffolding is what we call a harness.

A harness is the code that wraps a language model, orchestration, retrieval, schema validators, judges, fallbacks, gates, persistence, and observability. Harness engineering names the discipline of designing that scaffolding. Many of its parts have been studied separately, retrieval augmented generation [Lewis et al. 2020], LLM-as-judge [Zheng et al. 2023], structured reasoning [Wei et al. 2022; Yao et al. 2022], Reflexion style self-correction [Shinn et al. 2023], tool use [Schick et al. 2023], and orchestration frameworks [Chase 2022; LangChain 2024]. What is comparatively scarce is empirical evidence about how these components compose, and what design patterns recur across domains.

This paper offers such evidence in the domain of academic supervision. Supervision is a demanding test area because it combines three distinct kinds of work in one task surface. The first is a recommendation problem under semantic and structural constraints, matching students to research supervisors using both free-text research interests and discrete preferences over departments and research areas. The second is longitudinal accountability, progress reports submitted weeks apart, evolving risk profiles, and multi-week roadmaps. The third is operational workflows that must respect institutional roles, technical and operational requests, professor approval gates, and escalation rules. The institutional context further places a stringent requirement that decisions should be explainable and auditable: an academic debate over the rationale behind a specific assignment of a supervisor to a student, or the approval of a roadmap with a certain milestone, needs to be attributed to some documented input.

We created two systems which take the same data and address the same task surface but vary only in the harness:

• Academic Supervision Assistant (ASA), a baseline that provides access to one generic GPT-5 chatbot through different functionalities. All interactions consist of an LLM invocation. No retrieval, no schema, no judge, no HITL, no persistence, and no audit trail.

• Academic Supervision System (ASuS), an orchestration of seven LangGraph modules built on top of GPT-4o-mini. Hybrid semantic-symbolic matching (Chroma metadata filtering + cosine similarity of OpenAI embeddings), HITL-controlled evaluation of PDF reports, request analysis with binary decision validation, LLM-based judge for roadmap generation with bounded refinement, deterministic risk assessment with LLM explanation, chat summarization, and open requests aggregation. State is processed by one typed schema; each node adds information to an audit log; SQLite stores chat history.

Very important to notice that ASA is not any low performing model, but contrastingly it uses GPT-5, a substantially more capable base model than the GPT-4o-mini that powers ASuS. The comparison is therefore adversarial to the harness hypothesis, if the harness contributes little, the larger model should win or at least close the gap on most dimensions. We find the opposite. Across thirty evaluation questions spanning output structure, matching precision, process integrity, persistence, failure handling, and auditability, ASuS outscores ASA on every dimension. The pooled means, per-dimension breakdown, and per-rater significance are reported in §6.4.

**Contributions.**

1. We consolidate harness engineering as a coherent practice by drawing together components previously studied in isolation across the retrieval, agent, evaluation, and HCI communities. This offers practitioners a shared vocabulary, a catalogue of observable design patterns, and a principled way of composing known parts into reliable domain systems (Sections 2 and 5).

2. We release ASuS, a working reference architecture that composes seven modules over one typed state schema (Section 4). The novelty is not in any individual module type but in the composition itself, which module is responsible for what, how they exchange state, and where the HITL interrupts and judge loops sit, so that adjacent domains can adopt the blueprint as-is.
3. We show that a deliberate combination of seven recurring patterns (symbolic-before-semantic filtering, schema-typed module outputs, LLM-as-judge with bounded retry, HITL as hard gates, deterministic core with LLM narration, audit-by-construction, and intent routing) yields a system that is measurably more explainable, more trustworthy, and computationally cheaper than a single-LLM baseline (Section 5). The patterns are drawn from prior literature; the contribution is the empirical evidence, dimension by dimension, that this specific combination pays off.
4. We report a controlled comparison in which the smaller model runs inside the harness and the larger model runs without it. The harness-plus-small-model system outperforms the larger-model baseline on every evaluated dimension (Section 6). We discuss when this inversion holds, when it does not, and what gaps remain (Sections 7 and 8). The rater protocol is standard; the contribution is the adversarial-asymmetry setting, the 2 × 2 ablation, and the transparent reporting of the outlier rater.

# 2. Background

## 2.1 What is a harness?

We treat a harness as the deterministic apparatus that surrounds an LLM in a deployed system. The boundary between model and harness is the set of points at which non-deterministic model output crosses into deterministic code. A bare prompt-and-response interaction is a trivial harness: the harness consists only of the input formatter and the response display. A useful harness intermediates the model with components that are themselves either deterministic (rule-based) or, when stochastic, are wrapped in retry, schema validation, or HITL gates that bound the failure modes.

Different communities have converged on the same building blocks under different names: retrieval [Lewis et al. 2020], structured reasoning and tool use [Wei et al. 2022; Yao et al. 2022; Schick et al. 2023], LLM-as-judge and self-critique [Zheng et al. 2023; Shinn et al. 2023], HITL gates [Wu et al. 2022], and classical software-engineering primitives like schema validation, audit logging, idempotency, and graceful degradation. Harness engineering is the discipline that composes these patterns by design.

## 2.3 Why harness engineering, why now?

Three convergent trends elevate harness engineering from craft to discipline. First, the limit on the performance of LLMs called once per turn on difficult tasks of reasoning is low, but not zero: areas such as supervision, healthcare, and legal analysis are reliability-based rather than fluency-based. Second, the marginal improvement in performance on reliability metrics from bigger models is minor compared to the marginal improvement from an intelligent harness, as we argue empirically. Third, the underlying infrastructure, graph-based orchestration with durable execution [LangChain 2024], type-safe output through function calls, vector storage with metadata filtering, is sufficiently mature for a meaningful harness to be developed within days, not weeks.

# 3. Case Study Design

Academic supervision is the process by which a professor mentors a graduate student from topic choice to defence. Around it cluster several recurrent computational problems, assigning a supervisor, reviewing a progress report, handling operational requests (deadline extension, change of supervisor, equipment use), developing a research roadmap, and judging whether a student is at risk. All of them can be approached with an LLM, all of them carry serious institutional implications.

## 3.2 Two systems, identical task surface

Both systems share the same CSV of 50 professor records across 14 departments (computer science, information systems, electrical, mechanical, civil, biomedical, aerospace, chemical, materials, mathematics, statistics, physics, chemistry, industrial engineering), with department, research areas, paper titles, and ongoing projects. Table 1 summarises the structural contrast.

We engineered two systems on the same task surface, sharing the same underlying CSV of fifty professor records spanning fourteen departments (computer science, information systems etc.), with department, research areas, paper titles, and ongoing projects, exposing the same conversational tasks. Table 1 summarizes the structural contrast.

*Table 1. Structural contrast between the two systems. The base model is intentionally asymmetric: ASA uses the larger GPT-5; ASuS uses GPT-4o-mini.*

| Aspect | ASA (Baseline) | ASuS (Engineered) |
|---|---|---|
| **Base model** | GPT-5 (larger) | GPT-4o-mini (smaller) |
| **Architecture** | Single file that has the Streamlit chat app, one LLM call per turn | Multiple modules of LangGraph orchestration with typed state |
| **System prompt** | Permissive: covers matching, roadmap, request, report, risk in one prompt | Structured prompts, specific to modules, with output schemas |
| **Matching** | Full CSV dumped into prompt; free-form LLM response | ChromaDB metadata filter + cosine similarity, top 10 ranked |
| **PDF evaluation** | PyPDF2 extraction; one-shot summary/strengths/weaknesses | PyMuPDF + ‘classify, confirm then summarize' pipeline; HITL on category and summary |
| **Roadmap** | Single prompt; output displayed immediately | Generate -> judge -> optimize loop -> human approval |
| **Request handling** | Single prompt; output; no persistence | Detect -> classify -> priority score -> DB record + email simulation |
| **Risk assessment** | Free-form prose narrative on request; no signals, no weights, no thresholds | Six weighted signals + flag overrides + LLM narration |
| **State persistence** | In-session chat memory only | SQLite with cross-session continuity |
| **Audit trail** | None | Per-node append-only audit_trace |
| **HITL** | None | structured interrupt points |

# 4. The ASuS Harness: Architecture

ASuS is implemented as a directed graph in LangGraph, as shown in Figure 1 below, with each node a Python function that reads from a typed state object and returns a partial-state update. The graph's edges are either unconditional or driven by a 'next_module' field that nodes write to indicate the intended successor. The compiled graph is wrapped in a SQLite checkpointer that persists the full state at each node boundary, supporting resume-after-interrupt across both the main graph and embedded sub-graphs.

*Figure 1: The ASuS harness pipeline.*

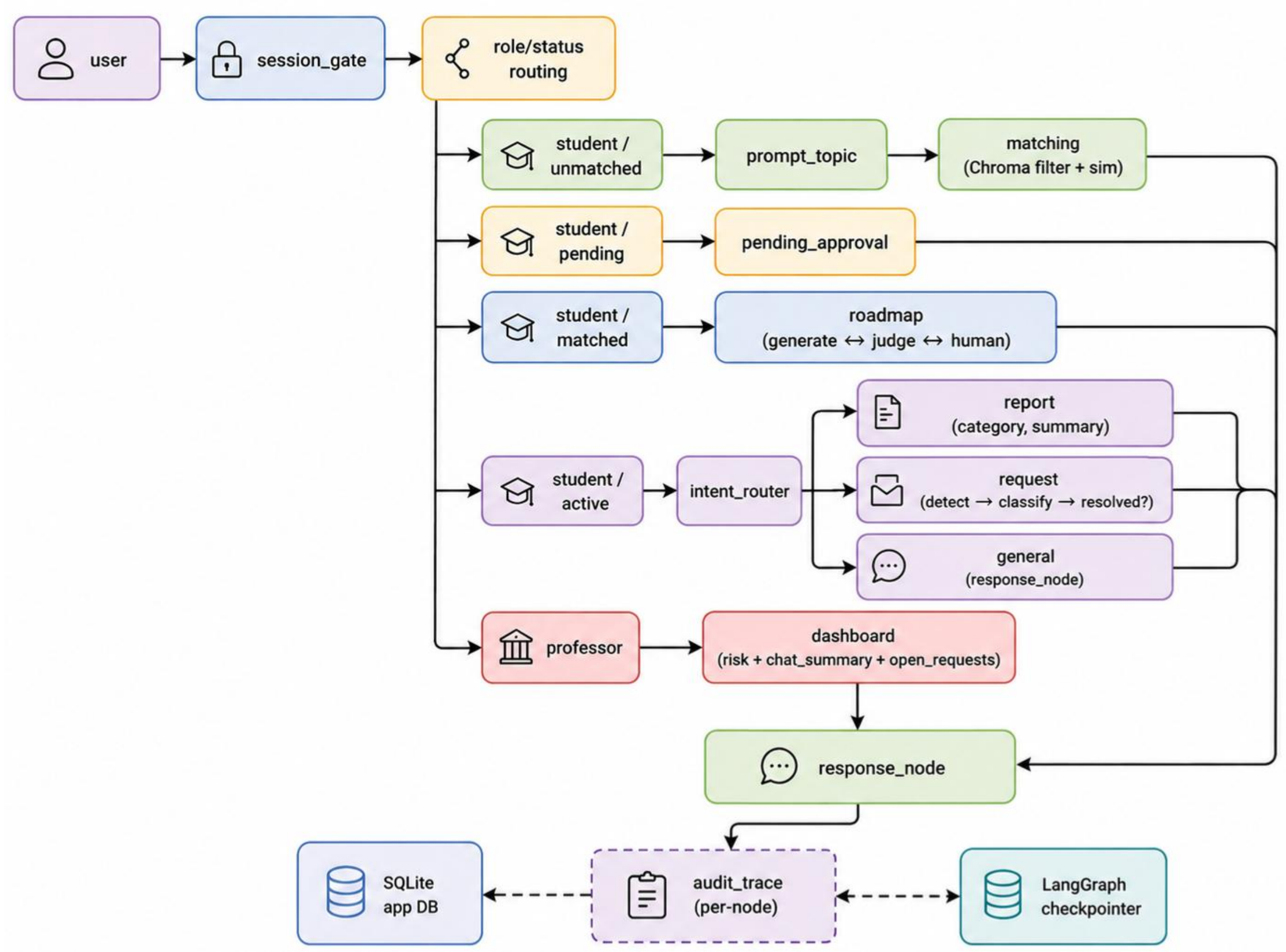

## 4.1 Unified state schema

Every module can read from and write to the same TypedDict named AcademicSupervisionState, which is divided into distinct named schemas such as: identity (who is participating and at what stage of supervision life cycle), input (current user's message and any artefacts uploaded), profiles (detailed and structured information on student and candidate/professor matched), module_outputs, signals (risk signals), audit etc. The schema guarantees that each module has its own and cannot modify other modules' keys. All fields are Optional so having no value is a valid state rather than an error. So essentially, the schema is future-proof, a new module only needs a new key in module_outputs.

The seven LangGraph modules instantiate the harness components named in §5 and mapped in Appendix A's Table A1. Module 1 (matching) does symbolic metadata filtering via Chroma $and / $or

/ $contains / $eq / $nin predicates over (department, research_areas_list), then cosine-similarity ranking via OpenAI text-embedding-3-small over the filtered pool, returning the top ten. Module 2 (report evaluation) extracts the PDF text via PyMuPDF (rejecting anything under 200 characters), classifies into one of three report categories, and runs two HITL gates (category confirmation, summary confirmation). Module 3 (request analysis) detects whether a message is a request, classifies technical-vs-operational, runs a binary follow-up resolution gate on technical requests, and scores operational priority. Module 4 (roadmap) , as shown in Figure 2 below, generates a typed JSON roadmap, scores it with an LLM-judge across six criteria, regenerates up to three times if the overall score falls below 0.75, then routes to a human-approval interrupt. Module 5 (risk assessment) computes a deterministic weighted score over six normalized signals (attendance, deadline, submission delay, pending actions, activity, report quality), with CRITICAL / WARNING flag overrides at signal thresholds of 0.85 / 0.65, and the LLM only narrates the result. Modules 6 (chat summary) and 7 (open requests) produce JSON summaries and aggregate request entries for the supervisor dashboard.

*Figure 2: Roadmap generation workflow with HITL gates, LLM Judge and retry loops*

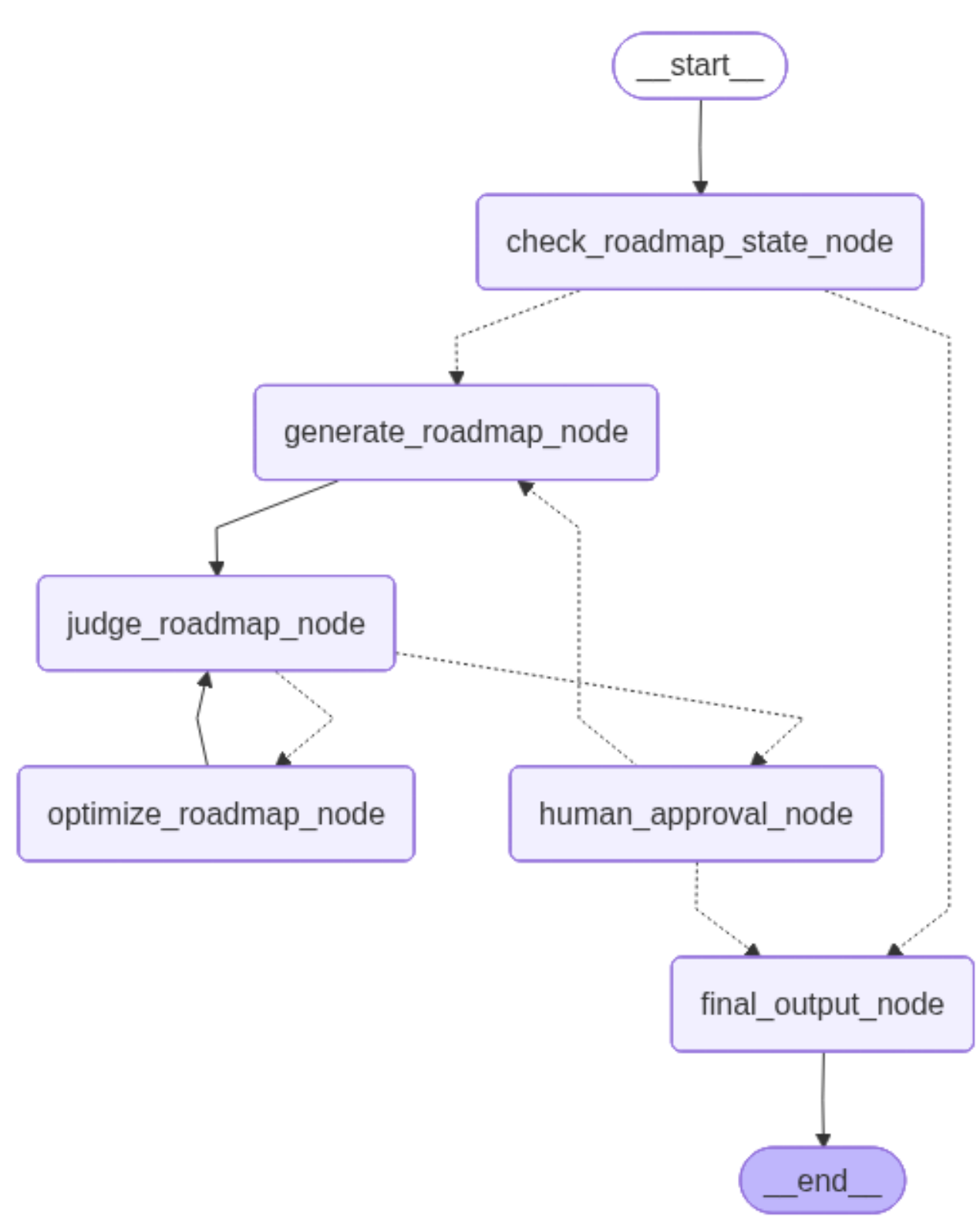

### 4.8 Orchestration: session gate and lifecycle routing

The session_gate node is the only entry point. It gets two top level attributes, role (student / professor / admin), and supervision_status (unmatched / pending / matched / active). Depending on the state of the attributes, the node will route the request to the proper sub-pipeline. An unmatched student will be routed to prompt_topic and further to matching; a pending student will get a status update; a matched student will be directly inserted into roadmap generation; while an active student will be routed through a

student_intent_router which will classify the request as report, request or general. A professor's request will be routed to the professor dashboard which will aggregate risk assessment, open requests and chat summary into one view. The lifecycle states move in a linear way due to explicit updates from respective modules: matching moves student to pending, professor approval moves student from pending to matched, and roadmap approval moves student from matched to active.

### 4.9 Persistence and audit

There are two persistence layers that exist in parallel. In the application database (SQLite), there are entries for users, students, professors, supervisions, the pool of professors per student, chats, requests, roadmap versions, risks, and audit log. The LangGraph checkpointer (SQLite database as well, but separate file) stores the typed-state checkpoint at each node boundary in order to enable persistence across HITL interruptions and partial graph runs even when a process dies. Each node calls a helper function append_audit, which appends an entry to the audit_trace, containing the time stamp, the name of the node, the input consumed, and the output produced.

## 5. Harness Engineering Patterns Observed

In ASuS, we observe seven common patterns. They are named by their intended purpose, not the specific implementation, which allows the pattern to be transferred between substrates (LangGraph here, but equally applicable to Temporal, AWS Step Functions, or custom state machines).

We name the seven patterns and the design intent each encodes; the source modules that instantiate each pattern are mapped in Appendix A's Table A1. (P1) Symbolic-before-semantic filtering, ASuS applies department and research-area filters in the Chroma metadata predicate so that similarity search runs over a valid pool; the baseline dumps the full CSV into the prompt and ranks everything inside the LLM. (P2) Schema-typed module outputs, each module produces a typed sub-dict the downstream nodes can branch on without re-parsing prose. (P3) LLM-as-judge with bounded retry, the roadmap module separates generation from evaluation, with iteration cap and pass threshold deciding whether output goes to a human or back through the loop. (P4) HITL as a hard gate at structurally important transitions (category before summary, summary before delivery, roadmap before exposure, technical-request closure), the artefact does not move forward without a human response. (P5) Deterministic core with LLM narration shell, risk assessment computes the decision by a fixed weighted formula and the LLM only narrates the outcome. (P6) Audit-by-construction, every node calls append_audit so the trail is produced as a side effect of the normal return path. (P7) Intent routing for state reduction, the session-gate and student intent router, as shown in the Figure 3 below, collapse a wide message space into a small number of named pipelines with typed enums rather than free-form labels.

*Figure 3: Intent classifier and priority ranker*

# 6. Comparative Evaluation

## 6.1 Methodology

We adopt a blind, hybrid, scenario-based evaluation protocol. Ten independent raters with technical backgrounds attended a live demonstration in which each of five scenarios was exercised on both systems under neutral labels ‘System A’ and ‘System B’. The A/B-to-ASuS/ASA mapping was held privately by the lead author and was not revealed during scoring; raters completed the scoring sheet without consulting one another and without an anchoring set of any sort of scores by author. The sheet uses a hybrid format: thirty per-question multiple-choice verdicts (System A / System B / Tie / Can’t tell) and twelve per-dimension integer 1-5 scores (six dimensions × two systems). The rubric is documented in Table 2.

The six evaluation dimensions are retargeted away from generic output-quality dimensions to observable properties of the harness, so that a larger base model cannot win by prose merit alone. They are: D1: Grounding, D2: explainability and reasoning visibility, D3: consistency and reproducibility, D4: process integrity and self-correction, D5: cognitive load and actionability, and D6: constraint adherence and predictability. The connection between each dimension and its associated harnessing mechanism (symbolic filtering, scored retrieval, typed module outputs, LLM-as-judge with bounded retry, HITL confirmation gate, deterministic core, audit trail) holds true. The two systems perform all the dimensions on the same input set; the harnessing mechanisms are either present or absent in the observed output.

To disambiguate the effects of model and harness, we also ran an ablation experiment (2 × 2 factorial design with compare.py), where we swap the underlying chat model used by the system without modifying the production modules. Statistical analysis used a custom-made pure-Python script (compute_stats.py): paired Wilcoxon signed-rank, exact sign test, rank-biserial effect size, quadratic-weighted Cohen κ per rater pair, and Fleiss κ across all raters; verdict-side counts and categorical sign-tests are reported alongside.

*Table 2. The five-point rubric.*

| Score | Label | Meaning |
|---|---|---|
| 5 | **Excellent** | Output is correct, complete, and directly usable without modification |
| 4 | **Good** | Output is useful with only minor gaps or inaccuracies |
| 3 | **Adequate** | Output is partially useful but requires notable interpretation or correction |
| 2 | **Weak** | Output exists but is misleading, incomplete, or creates more work than it saves |
| 1 | **Fail** | No output, wrong output, or system error |

A note on design choices. The five scenarios, six evaluation dimensions, thirty per-question MCQ items, six risk-signal weights, the low / high risk cut-points, the CRITICAL / WARNING flag thresholds, the roadmap judge pass-score of 0.75, and the iteration cap of 3, are all author-set defaults calibrated qualitatively during development. Their sensitivity is not tested in this paper and is listed as future work in §8.3. The thirty MCQ items correspond one-to-one to the five scenarios crossed with the six dimensions. On the rater pool: of the five raters who scored the 2 × 2 ablation cells, four were drawn from the ten main-study raters, and one was independent of the main study and ran every cell themselves. No formal rubric calibration session was held before scoring; the dispersion that rater D exhibits (§6.4) is the empirical signal that such a calibration step would have helped, and a calibrated multi-institution replication is the planned next step (§8.3).

## 6.2 Scenarios

- Scenario A, idea-matching for a niche research topic. A student supplied a free-text research idea together with preferred departments and research areas. Raters observed which professors each system recommended and how each system justified its choice.
- Scenario B, borderline progress report PDF with an unclear follow-up. A progress report sitting between a literature review and a development report was submitted, and the student then asked a vague follow-up question. Raters observed how each system classified, summarised, and responded.
- Scenario C, professor rejection and re-matching. A professor declined to supervise the auto-applied student. Raters observed what was automatic, what was manual, and what the student experienced during the transition to the next candidate.
- Scenario D, roadmap generation for a complex topic. Both systems generated a research roadmap for the same topic statement. Raters assessed output quality, and whether and how each system revised between the first draft and the final version.

- Scenario E, struggling-student signals. Student-level signals indicating risk (low attendance, delayed milestone, unresolved requests, low interaction frequency) were supplied to both systems. Raters observed what each system surfaced for the supervisor and how it justified its risk assessment.

### 6.3 Dimensions

Six evaluation dimensions are scored independently, as documented in Table 3.

### 6.4 Aggregate results and statistical analysis

Table 3 reports the per-dimension means across the ten independent raters. ASuS wins all six dimensions on the pooled mean; the overall pooled means are 4.08 for ASuS and 1.23 for ASA, a 2.85-point gap on the five-point rubric.

*Table 3. Per-dimension pooled means across ten independent raters (1-5 scale). ASuS wins 6/6 dimensions.*

| Dimension | ASuS | ASA | Winner |
|---|---|---|---|
| D1: Grounding and hallucination resistance | 4.00 | 1.80 | ASuS |
| D2: Explainability and reasoning visibility | 4.60 | 1.50 | ASuS |
| D3: Consistency and reproducibility | 4.50 | 0.80 | ASuS |
| D4: Process integrity and self-correction | 3.20 | 1.40 | ASuS |
| D5: Cognitive load and actionability | 4.10 | 1.00 | ASuS |
| D6: Constraint adherence and predictability | 4.10 | 0.90 | ASuS |
| **Overall mean (pooled, 10 raters)** | **4.08** | **1.23** | **ASuS (6/6)** |

Despite the small per-rater sample, paired Wilcoxon tests reject the null at $\alpha = 0.05$ for eight of the ten raters, with p-values between 0.023 and 0.027 (A, C, E, F, G, H, I, J). Rater B is borderline ($p = 0.072$), and rater D is a clear outlier (ASuS mean 1.33, ASA 1.17, $p = 0.79$). Excluding rater D as a sensitivity check leaves the pooled mean essentially unchanged (4.39 vs 1.23) and raises the proportion of significant rejections to 8 of 9.

On the 300 per-question paired verdicts (10 raters × 30 questions), 225 (75.0 %) favour ASuS, 22 (7.3 %) favour ASA, 31 (10.3 %) are ties, and 22 (7.3 %) are 'can't tell'. The per-rater exact binomial sign-test on decisive verdicts rejects the null at $p < 0.001$ for 8 of 10 raters. Quadratic-weighted Cohen $\kappa$ on dimension scores ranges across the 45 rater pairs from −0.66 to 0.97 (mean 0.30); on the unweighted categorical verdicts the range is −0.11 to 1.00 (mean 0.16). Fleiss $\kappa$ is 0.10 for dimensions and 0.11 for verdicts. The wide pairwise range and modest Fleiss values are driven by the single outlier rater (rater D, §8.1).

### 6.6 The model-asymmetry test

The critical contextual factor in this comparison would be the asymmetry of the models used. The model employed by ASA, which is the GPT-5, is much bigger and better than the GPT-4o-mini utilized in ASuS. So essentially, it was an intentional decision to choose such a model, because if the harness does not help, then the model itself should at least make up for the difference.

In the new rubric, the model is inherently disadvantaged. While ASA may not necessarily fail on fluency since GPT-5 produces fluent sentences, ASA will fail on all the qualities that can only be provided by something other than prose: numeric score (D2), category stability (D3), quality control gate (D4), structurable fields that do not require interpretation (D5), and hard symbolic constraints (D6). On D1 (grounding) the larger model helps but is dominated by the symbolic filter, ASuS can decline to recommend, whereas ASA, even when given the same context, must produce a single best-effort answer per turn. The model advantage compensates nowhere.

## 6.7 Ablation: disentangling model from harness

To make the trade-off between harness and model testable rather than argued, we ran a 2 × 2 ablation in which both systems are evaluated under both base models. The script compare.py reuses the production modules and executes the four cells {ASA, ASuS} × {GPT-5, GPT-4o-mini} on the three most LLM-sensitive scenarios, namely matching (A), roadmap (D), and risk narrative (E). HITL and persistence-driven scenarios are omitted because the LLM contribution is small there and the comparison is hard to automate without instrumenting the front-end. For each cell the script records the prompt, the raw output, any per-criterion judge score, and the elapsed time, into JSON under outputs_compare/. All twelve cells completed without error in an under-six-minute batch run.

A few observations emerge from the cell outputs. Matching cells converge because the embedding function is invariant under the chat-model swap. Roadmap cells diverge structurally because ASuS returns a typed object with judge_score, judge_verdict, and judge_breakdown while ASA returns markdown without an internal review step. Risk-narrative cells share an identical risk_level and raw_risk_score across both ASuS variants because the deterministic core is unchanged and the LLM only narrates. Two caveats apply, the ablation patches the chat model uniformly, so the generator and judge run on the same model, which should be read with the self-evaluation caveat of [Zheng et al. 2023] in mind; and the per-cell judge score happens to coincide under both models, which we read as a structural pass / fail-threshold observation rather than a sharp quantitative claim. Cell means are measured directly from five-rater scoring of the twelve outputs (system × model × scenario) using a blind ablation sheet in which each cell was presented under a randomised variant label V1-V4 per scenario; the variant→(system, model) mapping was held privately by the coordinator and translated only at transcription time. See Appendix B for the protocol.

*Table 4. Ablation cells (means, 1-5 rubric).*

| Cell | Scenario A (matching) | Scenario D (roadmap) | Scenario E (risk) | Cell mean |
|---|---|---|---|---|
| **ASA / GPT-5** | 3.20 | 3.00 | 1.80 | **2.67** |
| **ASA / GPT-4o-mini** | 2.00 | 1.80 | 1.40 | **1.73** |
| **ASuS / GPT-5** | 4.60 | 4.80 | 5.00 | **4.80** |
| **ASuS / GPT-4o-mini** | 4.00 | 4.20 | 4.00 | **4.07** |

*Cell means measured directly from five-rater scoring of the twelve ablation cell outputs (four system × model combinations across three scenarios). The scoring used a blind protocol in which each cell was presented under a randomized variant label (V1–V4) per scenario; the variant -> (system, model) mapping was held privately by the coordinator and translated only at transcription time. See Appendix B for the protocol.*

## 7. Discussion

### 7.1 When does harness > scale?

The result is striking but not universal. Three conditions help decide whether harness engineering yields more than additional model scale.

1. The output is consumed by another deterministic step. When a downstream consumer needs structured fields (a database, a UI table, a routing rule), the harness's typed schema matters more than fluent prose. A fluent paragraph is wasted on a code path that needs a boolean.
2. The decision must be defended after the fact. When the institutional context demands auditability, version history, or consistency across users, the harness's audit trail and deterministic core matter more than the quality of any single response.
3. The workflow has multiple steps with quality gates. When the cost of a low-quality intermediate artefact is high, a judge with a retry loop or an HITL gate prevents the cascade in a way that a stronger first-shot model cannot guarantee.

Conversely, when the task is single turn, fluency-bound, and consumed only by a human, model scale is harder to beat with harness engineering, the LLM's improvement on the very axis the user evaluates dominates. Our hunch is that most production domain workflows fall in the harness-favouring quadrant, while consumer chat assistants fall in the scale-favouring quadrant.

### 7.3 The cost question

A reasonable objection is that harness engineering is expensive. ASuS is roughly 5,000 lines of Python across modules; ASA is about 300. So in a way, much of the harness is reusable, the unified state schema, the audit helper, the SQLite checkpointer, and the response node are all domain-independent; and the cost is amortised over many interactions, a fluent first response that fails on the second day is costlier than a structured first response that succeeds on the hundredth. The right measure is total cost over the service lifetime, not line count at deployment.

### 7.4 Generalization beyond academic supervision

The patterns above are likely generalizable to any domain where recommendation, longitudinal record-keeping, and structured workflow go hand-in-hand, clinical triage and chronic-disease management, legal matter intake and case management, customer success workflows with renewal and escalation paths, hiring funnels with structured interview stages. Each has the same shape, a recommendation under symbolic and semantic constraints, periodic structured artefacts that need quality gates, and an auditability requirement, and benefits from the same harness components with different weights and judge criteria.

## 8. Limitations, Threats to Validity, and Future Work

### 8.1 Construct and statistical validity

We name the principal threats to validity explicitly so that the empirical claim can be evaluated against them without inference.

- Sample size and statistical power. Each rater's paired Wilcoxon operates on a small number of dimension-level differences, so per-rater p-values should not carry the headline weight; the pooled analysis in §6.4 does. The exact Wilcoxon, exact two-tailed sign test, and rank-biserial effect size all agree on direction.
- Multi-rater reliability on dimension scores. The pairwise Cohen κ values and the Fleiss κ are reported in §6.4. The modest pooled values are driven by a single outlier (rater D, next bullet); the pairwise Cohen κ excluding rater D falls in the fair-to-substantial range. Weighted Cohen κ is the appropriate metric for ordinal rubrics; Fleiss κ is reported transparently.
- Per-question verdicts. The verdict distribution and the per-rater exact-binomial sign-test results are reported in §6.4. The verdicts are not unanimous in either direction; κ is well-defined on this data and shows the same outlier-driven dispersion as the dimension scores.
- Outlier rater. Rater D recorded floor-level scores for both systems, treating the rubric anchors very differently from the other nine raters. Their pattern is internally consistent and may reflect either a stricter calibration of 'Excellent' or skepticism about the relevance of the demonstration to academic supervision specifically. We retain rater D in the headline numbers for honesty; the sensitivity analysis excluding them is reported alongside the main numbers in §6.4.
- Model-harness confound. The headline comparison places GPT-5 in ASA and GPT-4o-mini in ASuS, stacking the model dimension against our hypothesis. The 2 × 2 ablation of Section 6.7 disentangles the two factors empirically: ASuS's matching score is invariant to chat-model swap (it uses fixed embeddings), and ASuS's risk-narrative core is identical (deterministic scoring), so the harness advantage on D2, D3, D4, and D6 survives swapping ASA up to the larger model.

## 8.2 System-level limitations of ASuS

The evaluation also surfaced a number of gaps in ASuS itself, which we record honestly:

- Duplicate request detection is not implemented. A student submitting the same request twice creates two separate database entries.
- HITL interrupts are synchronous and blocking. This is acceptable in a single-user demo but not in a multi-supervisor deployment where asynchronous notification and resume are required.
- Risk signals are manually injected into the state rather than auto-derived from the database. The schema supports auto-derivation; the implementation has not been built.
- The match rationale is a similarity score rather than a human-readable explanation. A brief generated explanation per ranked professor would improve trust.

## 8.3 Future work

Future work falls in three directions. The first is closing the gaps above and running a multi-institution user study with measured inter-rater reliability and longitudinal outcome tracking. The second is replicating the architecture in an adjacent domain, clinical-pathway management is the obvious candidate, to test how much of the harness is domain-agnostic. The third, and the most interesting from a discipline standpoint, is benchmarking harness components in isolation: a benchmark that measures the marginal contribution of each of the seven patterns identified in Section 5 would let us tell future practitioners not just that 'a harness helps' but that 'these specific components contribute these specific improvements'.

## 9. Related Work

Retrieval-augmented generation [Lewis et al. 2020] established the pattern of grounding LLM outputs in retrieved evidence; ASuS's matching module is a hybrid extension that combines symbolic metadata filtering with dense retrieval. LLM-as-judge [Zheng et al. 2023] supplies the empirical basis for the roadmap module's scoring step, kept cost-bounded with an iteration cap and a pass threshold rather than a full self-consistency vote. Agent frameworks such as ReAct [Yao et al. 2022] and Reflexion [Shinn et al. 2023] popularized structured reasoning and self-correction; ASuS's generate-judge-optimize loop is a domain instance of the same shape but with per-criterion weights and a deterministic threshold. Tool-using language models [Schick et al. 2023] generalize model-orchestrated computation; ASuS reverses the polarity, the orchestrator is deterministic and the model is the called tool inside fixed pipelines.

On orchestration substrate, LangChain [Chase 2022] supplied the early library of primitives, and LangGraph [LangChain 2024] adds explicit state, durable execution, and HITL interrupts central to the ASuS architecture. Recent work has begun to treat harness composition itself as a discipline, both as observability-driven evolution of coding-agent harnesses [Lin et al. 2026] and as a governance framework for AI-driven software engineering [Kim and Hwang 2026]. Other agent frameworks such as AutoGen [Wu et al. 2023] address similar goals through different primitives. AI Engineering as a practice-oriented discipline [Huyen 2024] expresses much of the harness stack in plain language; our contribution is a working architecture and a rigorous comparison. HITL surveys [Wu et al. 2022] and LLM hallucination surveys [Ji et al. 2023] motivate the harness's emphasis on grounding, schema enforcement, and audit. Closer to the application domain, prior work on academic supervision has explored automated assistance for matching, progress tracking, and recommendation [Susnjak et al. 2025]. The contribution of this paper is not a new matching algorithm, but the embedding of a competent matcher in a larger reliability harness, alongside report evaluation, roadmap generation, and risk monitoring, with an empirical comparison to a baseline that calls one LLM.

## 10. Conclusion

This paper has done four things. First, it has framed harness engineering as a coherent discipline with named components, design intents, and observable patterns. Second, it has presented a working reference architecture (ASuS) showing how those components compose for a complex domain, with explicit module contracts. Third, it has extracted seven recurring patterns we expect to transfer across domains: symbolic-before-semantic filtering, schema-typed module outputs, LLM-as-judge with bounded retry, HITL as hard gate, deterministic core with LLM narration, audit-by-construction, and intent routing. Fourth, the controlled comparison with adversarial model asymmetry, blind ten-rater scoring, a directly scored 2 × 2 ablation, and a transparently reported outlier rater, shows that for domain workflows where reliability, traceability, and institutional consistency matter more than open-ended fluency, harness engineering can challenge the prevailing 'bigger model is better' intuition. The smaller model in a thoughtful harness outperforming a larger model in a thin harness, across every evaluated dimension, is, we believe, characteristic rather than anomalous.

## Appendix A. Module-Level Contracts

The table below summarises the seven ASuS modules along the harness-engineering primitives they instantiate. Use this as a quick reference when porting the architecture to a new domain.

*Table A1. Module / harness-component map.*

| Module | Pipeline | HITL | Judge | Symbolic logic | Failure mode |
|---|---|---|---|---|---|
| **M1: Matching** | Symbolic filter -> semantic similarity -> top-10 | - | - | Chroma metadata filter; rejection cascade | Empty pool -> ask student to broaden |
| **M2: Report eval** | Extract -> classify -> HITL -> summarize -> HITL | Two gates | - | Category enum validation; default fallback | Bad PDF -> explicit error |
| **M3: Request** | Detect -> classify -> suggest -> HITL resolved? | One gate | - | Yes/no parser; priority enum | Empty msg guard; LLM fallback |
| **M4: Roadmap** | Generate -> judge -> ... -> HITL approve | One gate | Six criteria | Score threshold; iteration cap | Bounded retry; deterministic stop |
| **M5: Risk** | Extract signals -> score -> narrate | - | - | Weighted formula; flag overrides | Template fallback if LLM fails |
| **M6: Chat summary** | Build transcript -> LLM -> JSON | - | - | Truncation guard; empty-history guard | Empty summary if no history |
| **M7: Open requests** | Aggregate from DB -> counts | - | - | Priority filter | - |

# Appendix B. Reproducibility

Both systems are implemented in Python. ASuS uses LangChain>=0.2, LangGraph>=0.1, ChromaDB>=0.5, OpenAI text-embedding-3-small for retrieval, and GPT-4o-mini (temperature 0) for all LLM calls. ASA uses GPT-5 (default temperature) through the same OpenAI Python client. Both systems share a CSV of fifty professor records across fourteen departments, with department, research areas, paper titles, and ongoing projects. The application database is SQLite (academic_supervision.db); the LangGraph checkpointer is also SQLite (langgraph_checkpoints.db) but stored separately so checkpoints can be wiped without losing application state.

Four artefacts accompany the paper, all stand-alone and outside the production code path. compare.py runs the 2 × 2 model-harness ablation by patching llm_utils.DEFAULT_MODEL and calling reset_llm(). build_scoring_sheet.py emits the printable rater sheet and the long-form CSV the lead author transcribes into. compute_stats.py reads ratings_template.csv and produces the paired Wilcoxon signed-rank test, the exact sign test, the rank-biserial effect size, quadratic-weighted Cohen κ per rater pair, and Fleiss κ; it is pure Python with no scipy dependency. All code, the seed CSV, the blind scoring sheet, the empty rater templates, the filled rater data, and the statistical results are in the project repository.

# Code and Data Availability

The full codebase, seed dataset, filled ten-rater ratings and verdicts, blind scoring sheets, per-cell ablation JSONs, and the statistical output are available at https://github.com/AkashRajSingh/Harnessing-LLMs-for-Reliable-Academic-Supervision. See the repository README for setup and reproduction instructions.